\documentclass[letterpaper, 10 pt, conference,  multi, tikz]{ieeeconf}  
\IEEEoverridecommandlockouts                              
\usepackage{graphicx}
\usepackage{subfig}
\graphicspath{{images/}}
\usepackage{times}    
\usepackage{svg}
\usepackage{amsmath}     
\usepackage{amssymb}     
\usepackage{graphicx}
\usepackage[noend]{algpseudocode}
\usepackage{url}
\usepackage{amsmath}
\usepackage[linesnumbered,ruled]{algorithm2e}
\usepackage{siunitx}
\usepackage{colortbl}
\let\labelindent\relax
\usepackage{enumitem}
\usepackage{cite}
\usepackage{float}

\usepackage[font=small]{caption} 
\usepackage{tabularx}
\usepackage{makecell}
\usepackage{multirow}
\usepackage{hhline}
\usepackage{tikz}
\usetikzlibrary{backgrounds, fit, shapes.geometric, positioning}
\usepackage[noend]{algpseudocode}
\usepackage{mathtools}
\usepackage{amssymb}

\newcolumntype{Y}{>{\centering\arraybackslash}X}
\usepackage[font=small]{caption}

\def\figref#1{Fig.~\ref{#1}}

\def\eqref#1{Eq.~(\ref{#1})}

\newcommand\etal{\emph{et al. }}

\newcommand*{\algotitle}[2]{%
  \stepcounter{algocf}%
  \hypertarget{algocf.title.\theHalgocf}{}%
  \NR@gettitle{#1}%
  \label{#2}%
  \addtocounter{algocf}{-1}%
}

\usepackage{makecell}
\usepackage{footmisc}
\usepackage{tabu}

\title{\Huge \bf Learning Goal-Oriented Non-Prehensile Pushing in Cluttered Scenes}

\author{Nils Dengler \and David Gro\ss klaus \and Maren Bennewitz
\thanks{All authors are with the Humanoid Robots Lab, University of Bonn,  Germany.
 		This work has partially been funded by the European Commission
  		under grant agreement number 964854 --RePAIR --
                H2020-FETOPEN-2018-2020 and
                by the Deutsche Forschungsgemeinschaft (DFG, German Research Foundation) under Germany's Excellence Strategy, EXC-2070 -- 390732324 -- Phenorob.}
  }

\begin{document}
\maketitle
\thispagestyle{empty} 
\pagestyle{empty}
\begin{abstract} 
Pushing objects through cluttered scenes is a challenging task,
especially when the objects to be pushed have initially unknown
dynamics and touching other entities has to be avoided to reduce the
risk of damage. In this paper, we approach this problem by applying
deep reinforcement learning to generate pushing actions for a robotic
manipulator acting on a planar surface where objects have to be pushed
to goal locations while avoiding other items in the same
workspace. With the latent space learned from a depth image of the
scene and other observations of the environment, such as contact
information between the end effector and the object as well as
distance to the goal, our framework is able to learn
contact-rich pushing actions that avoid
collisions with other objects. As the experimental results with a six
degrees of freedom robotic arm show, our system is able to
successfully push objects from start to end positions while avoiding
nearby objects. Furthermore, we evaluate our learned policy in
comparison to a state-of-the-art pushing controller for mobile robots
and show that our agent performs better in terms of success rate, collisions with other objects, and continuous object contact in various scenarios.
 \end{abstract}

\section{Introduction}
\label{sec:intro}
%

Pushing is often used for re-positioning and re-orientating objects since it simplifies the object manipulation in comparison to pick-and-place approaches. 
Furthermore, pushing allows for moving large, heavy, and irregularly shaped, as well as small and fragile objects to target positions and can be used for reducing uncertainty in the position of objects~\cite{mason1986mechanics}. 
Hereby, the term pushing is separated in non-prehensile pushing~\cite{lloyd2021goal} and prehensile pushing (push-grasp)~\cite{dogar2010push, bejjani2019learning}.
For example, in limited space\mbox{\cite{cosgun2011push,bejjani2021learning}} and when dealing with fragile objects, non-prehensile pushing is the preferred manipulation action, since grasping increases the risk of damage. 
In the past, pushing has been used to separate objects for better grasping~\cite{eitel2020learning, zeng2018learning} or to sort objects from a table into a bin~\cite{ewerton2021efficient} and is assumed to be more time-efficient than grasping to overcome short distances~\cite{push-Net}. 
The range of pushing actions vary between a few centimeters for corrective actions to larger distances, e.g, to place an object in the last row of a shelf.
Several approaches aim at predicting the physical properties of objects and use short pushing actions of predefined length \cite{paus2020predicting, nematollahi2020hindsight}.

\begin{figure}[ht!]  
\centering 
\includegraphics[width=0.8\linewidth]{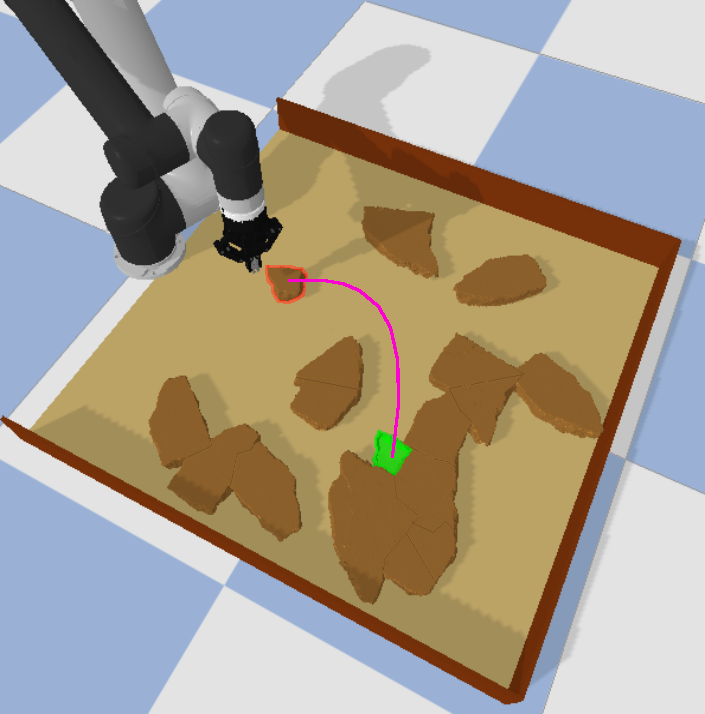}
\caption{Targeted application scenario of our system within
  the RePAIR-project\footref{repair}. The goal is to push the small
  fragment to the desired goal pose~(green). Shown in magenta is the
  best pushing path, which maintains a safety distance to the other
  objects.} \label{fig:cover}
\end{figure} 

In general, pushing actions should be contact-rich with smooth arm motions. Furthermore contact to other objects in the workspace should be avoided to prevent any damages and changes the configuration of the scene.
While for a long time, pushing behaviors were created using expert knowledge in an analytical way, more and more work is focusing on reinforcement learning~(RL) to solve this task. 
Especially the ability to learn from environment interactions and own experiences makes RL a useful way to learn challenging new skills.
Start-to-goal pushing with an RL-agent has been tackled
before~\cite{lin2019reinforcement} and serves as a benchmark for
RL~\cite{rl-zoo3}, however, pushing in cluttered environments where collisions with other objects have to be avoided is a less researched area. 
While there are already approaches for mobile bases~\cite{krivic2019pushing,stuber2018feature}, they have not been transferred to robotic manipulators so far.\\
\indent In this paper, we present a framework to train an RL-agent that is able to realize obstacle-aware pushing in a contact-rich manner to guide objects with initially unknown dynamics on a planar surface to desired target configurations with a robotic manipulator.
As representation of the workspace, we use a depth image taken from a bird's eye view. 
To reduce the size of the observation space and therefore the complexity, we use the latent space of a variational autoencoder. 
To accelerate learning, we calculate the optimal 2D~path in a grid representation of the environment generated from the depth image.
From this path we sample subgoals, which we use as observations to our agent.
In addition we use further observations, such as contact information between the end effector of the manipulator and the object as well as the distance to the goal. 
The output of our system is an incremental motion of the current $(x,y,\theta)$-position of the robot's end effector.
\figref{fig:cover} illustrates a targeted application scenario from the RePAIR-project\footnote{https://www.repairproject.eu \label{repair}}. The goal is to push the small fresco fragment to the desired position in a gentle manner while not damaging it or any other fragment on the assembly table.

The key contributions of our work are the following
\begin{itemize} 
\item A model-free RL system that learns to generate smooth pushing paths, with
  contact-rich pushing actions to reach the object's target positions in
  cluttered environments, thereby avoiding contact to other, nearby objects.
\item A qualitative and quantitative evaluation in simulation in
  comparison to a state-of-the-art pushing controller~\cite{krivic2019pushing}, which we adapted to our scenario.
\end{itemize} 
As the experiments with a six degrees of freedom robotic arm show, our system leads to reliable pushing, while achieving 
better performance compared to~\cite{krivic2019pushing} with respect to success rate, collisions with other objects, and continuous object contact in various scenarios. 

\section{Related Work}
\label{sec:related}
\begin{figure*}[ht!] \centering 
\includegraphics[width=0.9\textwidth]{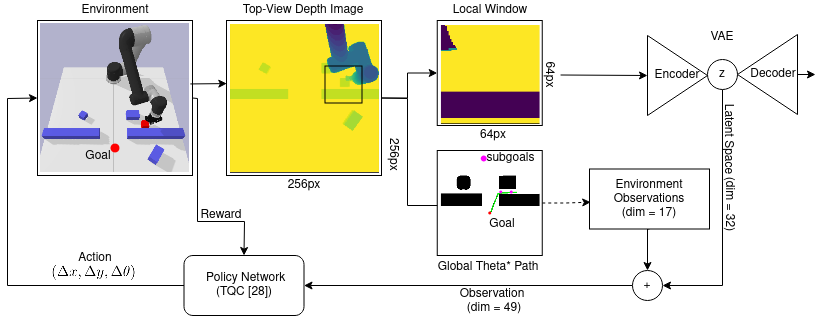}
\caption{Overview of our deep reinforcement learning pushing framework. 
Our system receives a depth image of the environment taken from an RGB-D camera.
We calculate an object centered egocentric local window and feed it into the variational auto encoder to get the latent space. 
Furthermore, a global path from the current object position to the goal position, including subgoals, is calculated.
The latent space, the subgoals, and further observations from the environments are used as the concatenated observation for the policy network of the deep RL.
The policy network calculates the best 3D incremental motion of the gripper from the observation and the reward it gets from the last environment interaction.} 
\label{overview}
\end{figure*}

Without an exact model of the object dynamics, it is hard to predict the
moving behavior of objects.  Recent work has shown that these dynamics
can be learned, e.g., \mbox{Paus~\etal\cite{paus2020predicting}} proposed a
system to predict outcomes of pushing actions and used a graph net
representing object relations as in- and output and trained their
network on two million synthetic samples. To address the problem that
typically a large amount of data is needed to train a network,
Nematollahi~\etal\cite{nematollahi2020hindsight} proposed an
unsupervised learning strategy and used a combination of an inverse
and a forward dynamics model.  A restriction of both approaches is
that they rely on discrete actions.  
The resulting pushing actions are typically not contact-rich and need
a high number of interactions to guide the object towards the goal.

an RL-learning approach that aims at performing $(x, y, z)$ pushing actions in a
continuous way was presented by \mbox{Xu~\etal\cite{xu2021cocoi}} who
developed an agent that pushes objects by incorporating their
physical properties and ensuring that the objects will not fall over,
e.g., in the case of an empty bottle. While the motion is stuttering,
due to the physical properties, the approach achieves a good success
rate in a clutter-free environment. A promising subdomain of
general pushing is planar pushing, where the push action is considered
only in 2D. \mbox{Bauza~\etal\cite{bauza2018data}} proposed a control
model that is learned from only a few data points.  The authors
evaluated their system against an analytical model on the task of
following a given object trajectory. As an extension to
this system, Hogan \etal \cite{hogan2018reactive} trained a classifier
that evaluates the current pushing behavior to guarantee a smooth
trajectory that is close to the given one. \mbox{Doshi~\etal \cite{doshi2020hybrid}}
proposed a controller that uses differential dynamic programming to
generate the motion model. In all three approaches, the end
  effector and the object are both tracked with a motion capture
tracking system and no further knowledge about the environment is
included. For pushing, the authors assume that each object has four
possible contact points.  The goal was to reach a desired goal
location in a predefined orientation with as few as possible contact
switches. The assumption of only four contact points constrains the
pushing actions and might not lead to the optimal solution.

In terms of goal-oriented pushing,
Bejjani~\etal~\cite{bejjani2019learning} proposed an approach to push
an object in cluttered environments towards a goal configuration
where clutter in the scene is intentionally pushed away to clear the
path. 
In contrast to that, we try to avoid the nearby objects as much as possible to avoid any damages. 
Furthermore, Migimatsu~\etal\cite{migimatsu2020object} designed a task
and motion planning system that do not use global but only relative coordinates to
determine the position of the end effector to the target, which
leads to higher robustness against changes. 
We also use only relative coordinates in our observation
space. Additionally, to improve the learning behavior of our agent, we
apply techniques proposed by Lee~\etal\cite{lee2020making} and
Lin~\etal\cite{lin2019reinforcement} who included force and touch
sensor measurements into the observation space to encourage safe
pushing actions and increase the convergence of the agent. In
particular, our framework also considers information about
the contact between the end effector and the object in the observation
space.

Further work towards goal-oriented pushing was proposed by
Lloyd~\etal\cite{lloyd2021goal} who used data from a depth camera as well as a tactile sensor.
While the authors considered no other objects in the scene, they achieved good results with their control-based approach which is also transferable to curved non-flat surfaces.
Krivic~\etal\cite{krivic2019pushing} developed a motion controller for
a mobile robot that enables reliable pushing of objects of different
shapes on the floor through cluttered environments.  In
this paper, we use \cite{krivic2019pushing} as baseline approach and
implemented a modified version for object pushing on a planar workspace with a robotic arm.

\section{Problem Description}\label{problem}

In this work we consider the following problem. In a tabletop environment, a robotic arm is supposed to move an object from its current position to a 2D goal configuration.
To achieve this, we consider the end effector (EE) of the arm moving in planar space $(x, y, \theta)$. 
The robotic arm can be of any degree of freedom (DOF).
In addition to the pushing object, there are other objects which need be
considered as obstacles and which might obstruct the direct path to the end configuration.
The obstacles have to be avoided by the EE and the object at all time.
The goal of the RL-agent is to determine the best incremental movement
$(\Delta x,\Delta y,\Delta\theta)$ of its EE position at each time
step, to move the object with the EE as fast, but also as safe as possible to the goal position while avoiding obstacles on the way.
An RGB-D camera is mounted centered above the scene in bird's eye view
to obtain observations of the objects in the workspace.

\section{Our Approach}
We apply deep reinforcement learning to solve the task described
above. This is motivated by the fact that we expect to obtain smoother
trajectories as we would get with a pure control-based approach. 
Especially for traversing narrow passages the lack of parameter tuning can be beneficial. 
We use a variational auto encoder~(VAE) to decouple the feature
extraction of the given depth image from the policy learning process~\cite{raffin2019decoupling}. \figref{overview} shows  an overview of our proposed system.
In the following, we describe the VAE as well as our RL framework in detail.

\subsection{Variational Autoencoder}
\label{sec:VAE}
First, we describe the preprocessing of the input data as well as the network architecture.
The networks are implemented and optimized using tensorflow \cite{tensorflow}.

\subsubsection{Preprocessing}
To sense the current world state, i.e, the position of the object and each obstacle, we use a depth image that is gathered from bird's eye view.
To focus on relevant information and ignore distant obstacles that do not influence the next best motion, we use a object centered local window, that is oriented towards the object's orientation, see Fig. \ref{overview}. 
We use a 64x64 pixel window from the original 256x256 image.
Since the object's and the arm's position are given as individual components in the observation space, we set the corresponding pixels to the background value. 
We use a convolutional VAE to encode the current normalized local window of the scene into the latent space.
We gathered 700k training images in simulation via a random RL-policy and train the network for 10 episodes with a batch size of 256. The dataset contains 10\% of blank images to also recognize if no obstacle is around the object.
\subsubsection{Network Architecture}

As network architecture  we use four convolution layers for the encoder and six deconvolution layers for the decoder.
All convolution layers are followed by a batch normalization layer and use the rectified linear unit function as the layer's activation function. 
For the encoder we use max pooling after the first and average pooling after the third layer.
The outputs are distributions to directly compute the loss function of our VAE without using any further metric. 
For the encoder, we use an independent normal distribution and for the decoder an independent Bernoulli distribution. 
We only use the encoder as part of the observation space, whereas the decoder is ignored.
In our experiments, we found that a latent space of 32
works best in terms of training time and feature representation. 

\subsection{Reinforcement Learning}
\label{sec:RL}

RL can be considered as a control problem that can be modeled as a partially observable Markov decision process~(POMDP).
This means, that the agent cannot determine its exact state~$s_t$ at time step $t$, but has to rely on the current observation~$o_t$ to get $s_t \approx o_t(s_{t-1},a_t)$ for the last state~$s_{t-1}$ and the current action $a_t$. 
In the end, the goal is to find a stochastic policy~$\pi(a_t|o_t)$ that maximizes the expected reward~$R$ for each episode, where $T$ is the number of time steps and $\gamma$ a discount factor.
\begin{center}
 \begin{equation}
	max \mathbb{E} \left( \sum^T_{t=0} \gamma^tR(s_t,a_t)  \right)
\end{equation}
 \end{center}
For the implementation, we followed some ideas proposed by Regier~\etal\cite{regier2020deep}, which proposed a
RL-framework to navigate in cluttered environments with a mobile robot.
In the following we define the action and observation space, the reward function, the used RL-algorithm, the experience replay buffer strategy, as well as the learning strategy.

\subsubsection{Action Space}
We steer the robot with point control. 
Therefore, the action space consists of the three values, $(\Delta x, \Delta y, \Delta \theta)$, which are the increment to the current $x$ and $y$ position, as well as the yaw angle $\theta$ of the gripper. We
set the maximum value of $(\Delta x, \Delta y, \Delta \theta)$ to the
maximum distance change possible in one predefined time window.
  
\subsubsection{Observation Space}
The observation space of our RL-agent consists of 49 values, as shown in Table \ref{observation}. 
The EE position is defined as $(x, y, yaw, pitch, roll)$ and given in relative coordinates towards the objects frame. 
To give the agent an indication of the best path, we include two subgoals, also in relative coordinates, into the observation that we calculate from the current shortest path. 
The shortest global path is calculated on a binary map, gathered from the depth image, where all obstacles are inflated according to the half of the object's diameter.
Note that the agent never receives the complete shortest global path in its observation and that the shortest path as well as the subgoals are re-calculated at each time step. 
Therefore, our agent is not constructed as a path following agent but learns the best pushing behavior during training.
For the calculation we chose a point after 20\% of the global path length of time step t as first subgoal and the subgoal of time step t-1 as second one. 
We use subgoals from two different time steps to give the agent an indication of the progress it made in any direction.
The Boolean value "contact with obstacle" indicates if the EE touches the object at each time step. 
During our experiments, we tested different sets of observations and found that the one in Table \ref{observation} leads to the best behavior.

\begin{table}
\centering
{\footnotesize
    \begin{tabular}{|c|c|c|}
\hline
\textbf{observation} & \textbf{size} \\ \hhline{|=|=|}
Local window latent space & 32 \\ \hhline{|-|-|}
EE position at t & 5 \\ \hhline{|-|-|-|}
6D joint angle poses & 6 \\ \hhline{|-|-|-|}
Sub-goal at t-1 & 2 \\ \hhline{|-|-|-|}
Sub-goal at t-5 & 2 \\ \hhline{|-|-|-|}
Contact with obstacle & 1 \\ \hhline{|-|-|-|}
Object to goal distance & 1 \\ \hhline{|-|-|-|}\hhline{|-|-|-|}
Overall: &  49 \\ \hhline{|-|-|-|}
\end{tabular}
}
\caption{Overview of the observation space.}
\label{observation}
\end{table}

\subsubsection{Reward}
Our reward function consists of following three components:
\begin{equation}
    r_{\mathit{dist}}= \hspace{1cm}
\begin{cases}
    50,& \text{if goal reached}\\
    -r_{\mathit{g\_dist}}-r_{\mathit{o\_dist}},& \text{otherwise}
\end{cases} \hspace{10000pt minus 1fil}
\end{equation}
\begin{equation}
    r_{\mathit{collison}}=  \hspace{0.5cm}
\begin{cases}
    -10,& \text{if object out of bounds}\\
    -5 & \text{if collision occurred}\\
\end{cases} \hspace{10000pt minus 1fil}
\end{equation}
\begin{equation}
    r_{\mathit{touch}}=  \hspace{0.75cm}
\begin{cases}
    r_{\mathit{o\_dist}},& \text{contact to object}\\
    0 & \text{otherwise}\\
\end{cases} \hspace{10000pt minus 1fil}
\end{equation}
The first equation encourages the agent toward a faster learning behavior. Therefore, it rewards the agent with a high positive value for accomplishing the task and penalizes higher distances between object and goal as well as object and EE. We use the global path length for $r_{\mathit{g\_dist}}$ and the Euclidean distance for $r_{\mathit{o\_dist}}$.
To ensure equal rewards through different start goal configurations, we normalize both distances by their initial distance and scale it between 0 and 1.
When neither the arm nor the object have moved during a time step and both are at their starting position, the outcome would be $ r_{\mathit{dist}}= -2$. 
$r_{\mathit{collison}}$ penalizes each collision of the object with clutter in the scene or if the object gets pushed out of the boundaries of the predefined workspace.
The last part of the reward $r_{\mathit{touch}}$ considers the suggestion of Lin \etal \cite{lin2019reinforcement} and indicates if the agent has contact with the object. 
Since we calculate the distance between the EE and the center of the object, a small distance value remains, even if the EE has contact to the object.
Therefore, we negate the $r_{\mathit{o\_dist}}$ penalty of $r_{\mathit{dist}}$ each time the EE has contact to the object, to encourage a contact-rich behavior.

Together all three parts form the reward function $r_{\mathit{total}}$ of our agent:
 \begin{equation}
    r_{\mathit{total}} = r_{\mathit{dist}} + r_{\mathit{collision}} + r_{\mathit{touch}}
\end{equation}

\begin{figure*}[ht!] \centering 
\subfloat[\label{eval_environments:a}]{\includegraphics[width=.15\textwidth]{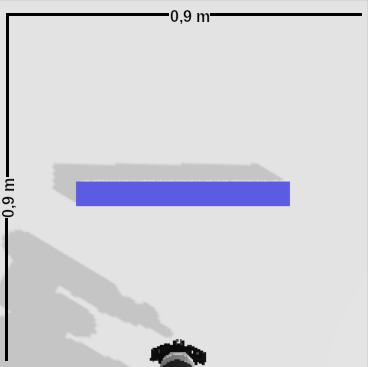}} \hspace{0.5em} 
\subfloat[\label{eval_environments:b}]{\includegraphics[width=.15\textwidth]{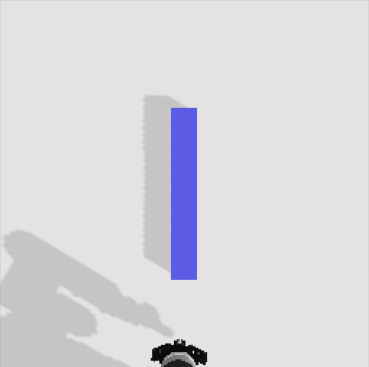}} \hspace{0.5em} 
\subfloat[\label{eval_environments:c}]{\includegraphics[width=.15\textwidth]{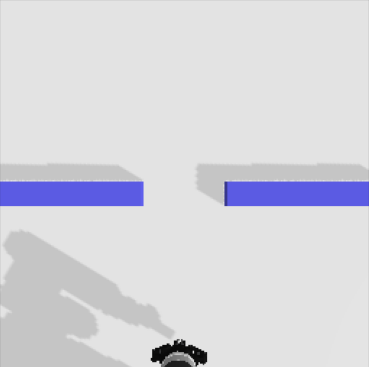}} \hspace{0.5em} 
\subfloat[\label{eval_environments:d}]{\includegraphics[width=.15\textwidth]{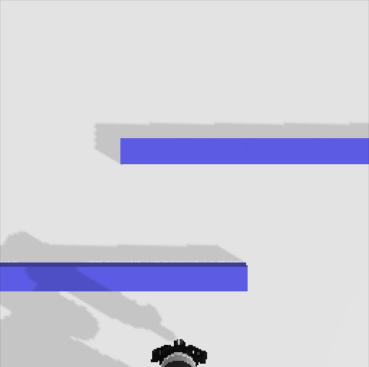}} \hspace{0.5em} 
\subfloat[\label{eval_environments:d}]{\includegraphics[width=.15\textwidth]{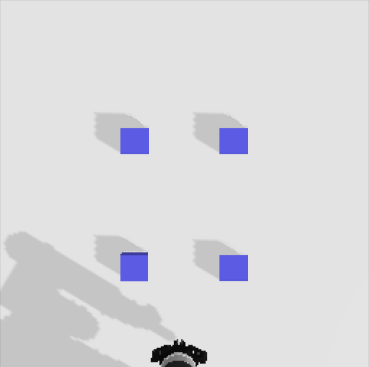}} \hspace{0.5em}
\subfloat[\label{eval_environments:a}]{\includegraphics[width=.15\textwidth]{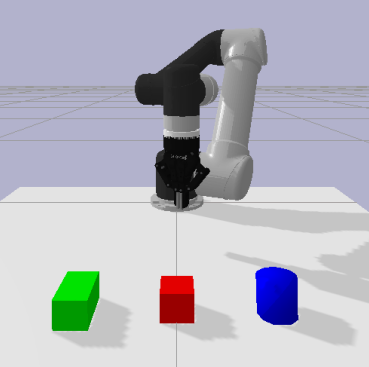}}
\hspace{0.5em} \caption{Figures (a) to (e) depict the different
  environments used for training and the quantitative evaluation. Figure~(f) shows
  the objects to be pushed. All objects have the same weight but
  differ in their geometrical shape. As pushing object during training we used the red cube. 
  In a curriculum learning manner, we rotated the obstacle in (a) and (b)
  and vary its size during the training. Furthermore, the distance
  between the obstacles in (c) to (e) decreased from 20 cm to  10 cm,
  making the task more difficult.} 
\label{eval_environments}
\end{figure*} 
\subsubsection{RL-Algorithm}
In this work we use the off-policy algorithm Truncated Quantile Critics (TQC) \cite{kuznetsov2020controlling}. 
During our experiment, TQC led to the best and the most reproducible results. 
The idea behind TQC is to control the overestimation bias in the critic's value estimation by using distributional critics \cite{kuznetsov2020controlling}. 
With multiple critics, the points of each of the distributions are used to create a mixture model to increase performance and stability. 
By truncating the last $n$ points from the mixture of distributions, the overestimation is alleviated.
In our work, we use the stable-baselines3 implementation of TQC \cite{stable-baselines3}.

\subsubsection{Attentive Experience Replay}
The experience replay strategy enables agents to learn from previous experiences they made while interacting with the environment. 
That means the agent stores action-state transition pairs in a buffer~$B$ of size~$N$ to reuse previous transitions to update the current policy. 
In case of the TQC algorithm with a sampling batch size~$\mathit{bs}$ the agent uniformly samples~$bs$ entries from~$B$ and reuses them. 
While the policy is evolving, some states are more frequently visited than others. 
For this reason, it is not useful to sample uniformly, because transitions that are rarely visited have a smaller benefit to the update process of the policy than frequently visited ones.
Therefore, Sun~\etal~\cite{sun2020attentive} proposed a new strategy to sample an entry from~$B$. 
With Attentive Experience Replay (AER) they suggest to sample entries according to the similarities between the entry's state and the current state of the agent. 
This means that according to the AER strategy, we uniformly sample $k \cdot bs$ entries from~$B$. 
We then calculate the similarity of each entry to the current state of the agent and use the $bs$~entries with the highest similarity score to update the policy. 
We use cosine similarity as the similarity measurement, a size of
$1e6$ for $B$, as well as $bs=512$ and $k=4$.

\subsubsection{Learning Strategy}
As the agent's learning strategy, we chose curriculum learning, which divides the task into subtasks and learns the subtasks one after another in increasing difficulty.  
We began the training with a maximum start-goal Euclidean distance of
0.06 m and increase it during training to up to 0.6 m. 
As training environments, we used the scenes shown in
\figref{eval_environments}.
The agent was trained for $7e6$ iterations.
Without curriculum learning the agent was not able to learn the task. 

\section{Experiments}
\label{sec:exp}
The goal of our experiments is to demonstrate the performance of our
system qualitatively and quantitatively in free space as well as in
obstacle-laden environments in terms of success rate, object contact,
number of collisions, and shortest path deviation, i.e.,
normalized inverse path length (SPL)~\cite{anderson2018evaluation}.
Furthermore, we provide a comparative evaluation against a
state-of-the-art pushing control approach by
\mbox{Krivic~\etal~\cite{krivic2019pushing}}. We performed the
evaluation in pybullet~\cite{coumans2021} with a 6 DOF UR5\footnote{https://www.universal-robots.com/products/ur5-robot/} with a Robotiq~2f85 two-finger gripper\footnote{https://robotiq.com/products/2f85-140-adaptive-robot-gripper}. 
We trained and evaluated our approach on a computer with an \mbox{i7-6800K} six-core CPU at 3.40 GHz and an Nvidia 2070 GPU with \mbox{8 GB} of memory used for the VAE. 
For actor and critic, we used a small network with three dense hidden layers of size [512, 256, 128].
For generalization, we used a Gaussian action noise with a standard deviation of 0.4.
As global 2D~path planner, to sample the sub-goals for the observation space, we used Lazy Theta* \cite{nash2010lazy}.
The implementation of our learning framework with all hyperparameters
as well as the reimplementation of the baseline approach is available at GitHub\footnote{https://github.com/NilsDengler/cluttered-pushing \label{github_code}}.

\begin{table} \centering 
\resizebox{.49\textwidth}{!}{%
\def\arraystretch{4}
\begin{tabular}{|c|c|c|c|c|c|} \hline 
\rowcolor[gray]{.75} {\Huge\textbf{Small Cube}}  & {\Huge\textbf{Success Rate}} & {\Huge\textbf{Object Contact Rate$\ast$}}& {\Huge\textbf{SPL}}& {\Huge\textbf{Path Length}}\\ 
{\Huge Ours} & {\Huge\textbf{1.000}} & {\Huge\textbf{0.943} $\pm$ 0.13}& {\Huge\textbf{0.928}}& {\Huge\textbf{0.429}$\pm$ 0.11}\\ \hline 
{\Huge Krivic \etal \cite{krivic2019pushing}}&{\Huge 0.998 }& {\Huge 0.870$\pm$ 0.15}&{\Huge 0.918}&{\Huge 0.430$\pm$ 0.10}\\ 
 \hline \end{tabular}} 
\caption{Quantitative evaluation of straight-line pushing in free
  space wrt. success rate, object contact, normalized inverse path length (SPL), and path length in meters.
The values are the average over 500~runs.
The results are in comparison to the approach by \mbox{Krivic~\etal~\cite{krivic2019pushing}} where the metrics marked with~a~$\ast$ are significant according to the paired t-test with a chosen p-value of 0.05. As shown, our approach performs better in terms of object contact and SPL and equally in terms of success rate.
}
\label{quant_res1} 
\end{table}

\subsection{Baseline Approach}
To compare our approach to the state of the art for pushing in cluttered environments, we reimplemented the controller proposed by Krivic \etal \cite{krivic2019pushing}. 
We implemented the approach as suggested in the paper and used the proposed parameters. 
Since the original approach was designed for a mobile base and with the assumption that the robot is always facing the pushing direction, we adapted our implementation to a robotic arm that starts at a position sampled around the object. 
Due to the sampling, the arm initially dragged the object with it when trying to reposition itself, causing unwanted collisions with obstacles or the object itself.
We adjusted this behavior by inverting the pushing direction, once the relocation activation of \cite{krivic2019pushing} surpasses a certain threshold $\Psi_{relocate}$.
This adjustment enabled the arm to reposition itself more efficiently.
The threshold for activating the inversion of the pushing direction was set to $\Psi_{relocate}>=0.6$.

\subsection{Quantitative Evaluation}
The quantitative evaluation consists of three parts, i.e.,
pushing in free space, in scenes with obstacles, and in previously unseen, highly cluttered scenes.
All metrics except the success rate and the SPL are evaluated only on episodes that both methods could solve successfully.
The object contact rate is evaluated for each episode, once the EE first touched the object.
Both, object contact rate and collision rate are the average of each episode, averaged over all episodes.
For all experiments, we randomly sampled the distance between start and goal within 0.2 to 0.6 m. 
As pushing object during training we used the red object shown in Fig.~\ref{eval_environments}.

\subsubsection{Straight-Line Pushing in Free Space}
We first evaluate straight-line pushing in scenes without other
objects to demonstrate the general pushing ability of the two
approaches.  
Therefore, we generated 500 start-goal configurations and compared the results to the shortest path found by Lazy Theta*~\cite{nash2010lazy}.
As shown in Tab.~\ref{quant_res1}, our approach performs slightly better in each metric
Especially, the higher object contact rate shows the impact of our reward function
guiding the agent towards the desired contact-rich pushing behavior.   

\begin{table}[] \centering 
\resizebox{.49\textwidth}{!}{%
\def\arraystretch{4}
\begin{tabular}{|c|c|c|c|c|c|} \hline 
\rowcolor[gray]{.75}  {\Huge\textbf{Small Cube}}  &  {\Huge\textbf{Success Rate}} &  {\Huge\textbf{Object Contact Rate $\ast$}}&  {\Huge\textbf{Collision Rate}} &  {\Huge\textbf{SPL}}&  {\Huge\textbf{Path Length $\ast$}}\\ \hline 
{\Huge Ours} & {\Huge\textbf{0.980}}& {\Huge\textbf{0.995} $\pm$ 0.02} & {\Huge\textbf{0.008} $\pm$ 0.04}& {\Huge 0.910} & {\Huge 0.523 $\pm$ 0.18}\\ \hline 
{\Huge Krivic \etal \cite{krivic2019pushing}}&{\Huge 0.955}& {\Huge 0.850 $\pm$ 0.10}&{\Huge 0.011 $\pm$ 0.05} &{\Huge \textbf{0.952}} & {\Huge \textbf{0.513}$\pm$ 0.16}\\ \hline  \end{tabular}}
\\ \vspace{1em}

\resizebox{.49\textwidth}{!}{%
\def\arraystretch{4}
\begin{tabular}{|c|c|c|c|c|c|} \hline 
\rowcolor[gray]{.75} {\Huge\textbf{Large Cube}}  & {\Huge\textbf{Success Rate}} & {\Huge\textbf{Object Contact Rate $\ast$}}& {\Huge\textbf{Collision Rate $\ast$}} & {\Huge\textbf{SPL}}& {\Huge\textbf{Path Length $\ast$}}\\ \hline 
{\Huge Ours} &{\Huge\textbf{0.977}}& {\Huge\textbf{0.995} $\pm$ 0.02}& {\Huge\textbf{0.007} $\pm$ 0.04}& {\Huge 0.910} & {\Huge 0.520$\pm$ 0.18}\\ \hline  
{\Huge Krivic \etal \cite{krivic2019pushing}}&{\Huge 0.957}&{\Huge 0.851 $\pm$ 0.10}&{\Huge 0.012 $\pm$ 0.05} &{\Huge \textbf{0.954}}& {\Huge\textbf{0.510} $\pm$ 0.16}\\ \hline  \end{tabular}}
\\ \vspace{1em}

\resizebox{.49\textwidth}{!}{%
\def\arraystretch{4}
\begin{tabular}{|c|c|c|c|c|c|} \hline 
\rowcolor[gray]{.75} {\Huge\textbf{Small Cylinder}}  & {\Huge\textbf{Success Rate}} & {\Huge\textbf{Object Contact Rate $\ast$}}& {\Huge\textbf{Collision Rate $\ast$}} & {\Huge\textbf{SPL}}& {\Huge\textbf{Path Length $\ast$}}\\ \hline 
{\Huge Ours} & {\Huge\textbf{0.967}}& {\Huge\textbf{0.981} $\pm$ 0.05}&{\Huge 0.021 $\pm$ 0.06} &{\Huge 0.839}  &{\Huge 0.553 $\pm$ 0.19}\\ \hline 
{\Huge Krivic \etal \cite{krivic2019pushing}}&{\Huge 0.945}&{\Huge 0.889 $\pm$ 0.11}&{\Huge \textbf{0.014} $\pm$ 0.04}&{\Huge \textbf{0.940}}&{\Huge \textbf{0.512} $\pm$ 0.17}\\ \hline  \end{tabular}}
\\ \vspace{1em}
 
\resizebox{.49\textwidth}{!}{%
\def\arraystretch{4}
\begin{tabular}{|c|c|c|c|c|c|} \hline 
\rowcolor[gray]{.75} {\Huge\textbf{Fragment}}  & {\Huge\textbf{Success Rate}} & {\Huge\textbf{Object Contact Rate} $\ast$}& {\Huge\textbf{Collision Rate} $\ast$} &{\Huge\textbf{SPL}}&{\Huge \textbf{Path Length} $\ast$}\\ \hline 
{\Huge Ours} & {\Huge 0.867}& {\Huge 0.980 $\pm$ 0.05}&{\Huge 0.05 $\pm$ 0.11}&{\Huge 0.71}&{\Huge 0.630$\pm$ 0.29} \\ \hline
{\Huge Ours re-trained} & {\Huge \textbf{0.959}}& {\Huge\textbf{0.984} $\pm$ 0.05}&{\Huge \textbf{0.01} $\pm$ 0.05}&{\Huge 0.83}&{\Huge 0.58$\pm$ 0.23} \\ \hline  
{\Huge Krivic \etal \cite{krivic2019pushing}}&{\Huge 0.953}&{\Huge 0.868$\pm$ 0.11}&{\Huge 0.024 $\pm$ 0.07 }& {\Huge\textbf{0.951}}& {\Huge\textbf{0.501} $\pm$ 0.16}\\ 
 \hline \end{tabular}}
\caption{Quantitative evaluation wrt. success rate, object contact, collisions, normalized inverse path length (SPL), and path length in meters. 
The values are the average over 1,000 runs.
The results are in comparison to the approach by Krivic \etal \cite{krivic2019pushing} where the metrics marked with~a~$\ast$ are significant according to the paired t-test with a chosen p-value of 0.05.
As shown, our approach achieves overall better results in terms of success rate, object
contact rate, and collision rate. Please refer to the text for more details.} 
\label{quant_res2}
\end{table} 
\begin{figure}[h] \centering 
\subfloat[Ours \label{hard_tasks:a}]{\includegraphics[width=.48\linewidth]{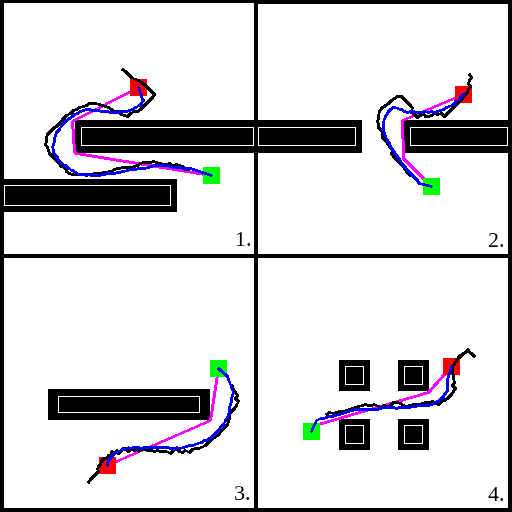}} \hspace{0.5em} 
\subfloat[Baseline \cite{krivic2019pushing}\label{hard_tasks:b}]{\includegraphics[width=.48\linewidth]{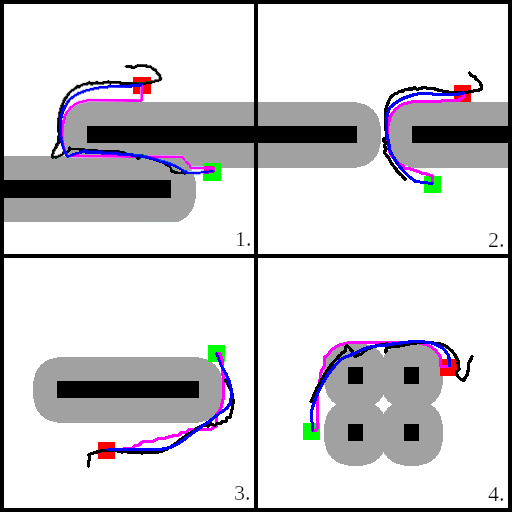}} \hspace{0.5em} 
\caption{Qualitative results from the quantitative evaluation of our
  approach (a) in comparison to the baseline~\cite{krivic2019pushing}
  (b). Red indicates the start, green the
  goal position and magenta is the initial shortest path calculated by
  Lazy Theta* \cite{nash2010lazy}. The path taken by the end effector  is shown
  in black and the path of the object in blue. The grey area in (b)
  shows the increased traversal costs around obstacles, used for the
  baseline approach, while the obstacles in our approach (a) are
  inflated only by a small amount according to the half of the object's diameter. As
can be seen, our agent learned to navigate around objects in a safe
  distance without strictly following the initial shortest
  path. Example~4 shows a result where our agent pushes a more
  efficient path, since it does not rely on any cost map.} 
\label{quant1_pic}
\end{figure} 
\subsubsection{Pushing in Scenes With Obstacles}
Furthermore, we generated five environments which differ in their
complexity, as shown in Fig \ref{eval_environments} (a) to (e).
We used three different types of objects, which were also used in~\cite{krivic2019pushing}, together with the completely unknown complex fragment object shown in \figref{fig:cover}, to demonstrate the generalization capabilities.
We sampled the orientation and size of the obstacles in (a) and (b) as well as the distance between the obstacles in (c) to (e).  In the following we refer to the
objects as "small cube"~(red), "large cube"~(green), and "small
cylinder"~(blue). 
For each object, we randomly generated 1,000 start-goal configurations
within the five environments.  As shown in Tab.~\ref{quant_res2}, our
approach again achieves a significantly higher object contact rate in
comparison to the baseline, which shows
the benefit of our approach in terms of gentle pushing through
contact-rich behavior.  Especially for scenarios as in the
RePAIR-project\footref{repair} gentile,
non-abrupt motions are crucial for not damaging any highly fragile
objects in the scene. 
Note that we used different objects for pushing, which the agent never
experienced during training. Still, the success rate is consistently high, except for the fragment where our agent still achieves a high success rate, without knowing any dynamics beforehand.
In terms of the SPL, the baseline achieves better
results while there is no significantly increased path
length. This behavior can be
explained with the higher obstacle inflation necessary for the
baseline approach and is illustrated in Fig.~\ref{quant1_pic} that
depicts example trajectories of the experiments.  As can be seen, our
agent has learned to safely navigate around objects, without strictly
following the initial shortest path.  This is a key advantage in
comparison to the baseline approach, which follows the shortest path
as tight as possible due to the properties of the controller method
and is crucial if the parameters are not fine-tuned.  This explains
the lower SPL of our approach. Note that with an increased inflation,
similar to the baseline, our SPL results will also increase.
Example~4 of~\figref{quant1_pic} shows a scenario where our agent
pushes a more efficient path, since it does not rely on any cost map and therefore on no parameter tuning.
As the fragment was never seen during training, we retrained the agent and achieved better overall results as without. This underlines, that our system
can be used for serving a general purpose but also retrained to specify on given scenarios. \\
\indent   For all experiments our policy
network took on average 0.791 ms for an action prediction of the
network and the simulation took 21.91 ms for realizing the action.
While our framework has a constant runtime, the runtime of the
baseline varies depending on the number of obstacles in the
environment and the corresponding greater computational effort.

\subsection{Pushing in Unseen, Complex Environments } 
\begin{figure} \centering 
\subfloat[\label{hard_tasks:a}]{\includegraphics[width=.45\linewidth]{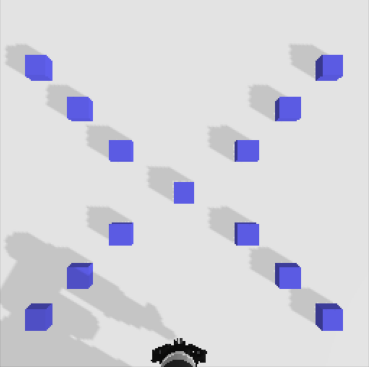}} \hspace{0.5em} 
\subfloat[\label{hard_tasks:b}]{\includegraphics[width=.45\linewidth]{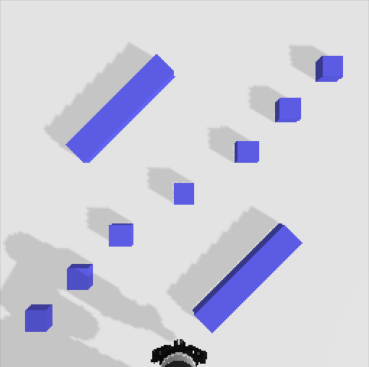}} \hspace{0.5em} 
\caption{Unseen, complex environments to further evaluate the
  performance of our system.} 
\label{hard_tasks}
\end{figure}
\begin{table} \centering 
\resizebox{\linewidth}{!}{%
\def\arraystretch{4}
\begin{tabular}{|c|c|c|c|c|c|} \hline 
\rowcolor[gray]{.75} {\Huge\textbf{Small Cube}}  & {\Huge\textbf{Success Rate}} & {\Huge\textbf{Object Contact Rate $\ast$}}& {\Huge\textbf{Collision Rate $\ast$}} & {\Huge\textbf{SPL}}& {\Huge\textbf{Path Length}}\\  \hline 
{\Huge Ours} & {\Huge\textbf{0.88}} & {\Huge\textbf{0.977} $\pm$ 0.06}& {\Huge 0.065 $\pm$ 0.13}& {\Huge\textbf{0.779}} & {\Huge\textbf{0.492}$\pm$ 0.13}\\ \hline 
{\Huge Krivic \etal \cite{krivic2019pushing}}& {\Huge 0.72}& {\Huge 0.566 $\pm$ 0.11}& {\Huge\textbf{0.01} $\pm$ 0.05}&{\Huge 0.720} & {\Huge 0.550 $\pm$ 0.18}\\ 
 \hline \end{tabular}}
\caption{Quantitative evaluation  in unseen environments with a high
  density of clutter~(\figref{hard_tasks}) wrt. the success rate,
  object contact rate, collision rate, the normalized inverse path
  length~(SPL) and the path length in meters in comparison to Krivic \etal
\cite{krivic2019pushing}.
The values are the average over 50~runs.
The results of the metrics marked with~a~$\ast$ are
significant according to the paired t-test with a chosen p-value of
0.05. As can be seen, our approach performs better in each metric
except the collision rate.} 
\label{qual_res1}
\end{table}

Finally, we designed more complex tasks with
the goal to evaluate the capabilities of our trained agent in unseen environments with a higher density of clutter.
We randomly sampled 50 start-goal configurations of the two
scenarios~(\figref{hard_tasks}), which contain many narrow passages. 
The results in Tab.~\ref{qual_res1} show the good performance in complex and completely unseen environments.
Our agent achieved better results than Krivic \etal
\cite{krivic2019pushing} in each metric except the collision
rate.
Especially the contact rate is significantly increased.
As already mentioned, our agent has not been trained on such
scenarios, accordingly, the success rate is a bit lower in comparison to the other evaluations with the small cube. 
\mbox{Regier \etal~\cite{regier2020deep}} showed that the success rate will highly increase while the collision rate will decrease, when the agent continues training in the unknown environment for a short time period.


%
%


\section{Conclusion}
\label{sec:conclusion}
In this paper, we presented a novel deep reinforcement learning approach for object pushing in
cluttered tabletop environments.
We demonstrated the efficacy of our approach in multiple simulated experiments where the results show the increased performance in comparison to an existing control-based method with respect to various metrics.
Our agent is able to perform pushing in free space and complex cluttered environments.
We showed that the pushing behavior highly benefits from our learning
approach in terms of constant object contact and smooth trajectories
avoiding obstacles while maintaining equal path length in comparison to
the  baseline method~\cite{krivic2019pushing}.
The evaluation of the runtime highlights that our system is capable of online pushing.
The code of our system can be found on Github\footref{github_code} and a video on our web page\footnote{https://www.hrl.uni-bonn.de/publications/dengler22iros.mp4}.

%


\bibliographystyle{IEEEtran}

\bibliography{references}

\end{document}